\documentclass[11pt]{article}

\usepackage{dialogue}

\usepackage{ifxetex}
\ifxetex
    \usepackage{fontspec}
    \setromanfont{Times New Roman}
\else
  \usepackage{cmap}
  \usepackage[T2A,T1]{fontenc}
  \usepackage[utf8]{inputenc}
  \usepackage{times}
  \usepackage{latexsym}
  \usepackage{substitutefont}
  \substitutefont{T2A}{\familydefault}{cmr}
\fi

\usepackage[russian,british]{babel}
\usepackage{url}
\usepackage{pgf}
\usepackage{array}
\usepackage{booktabs}
\usepackage{multirow}
\usepackage{tabularx}
\usepackage{makecell}
\usepackage[flushleft]{threeparttable}
\usepackage{ragged2e}

\usepackage{covington} 

\usepackage{hyperref}

\dialogfinalcopy 

\title{RuNNE-2022 Shared Task:\\
Recognizing Nested Named Entities }

\author{Ekaterina Artemova${}^{1,2}$, 
Maxim Zmeev${}^{1}$,
Natalia Loukachevitch${}^{3}$, 
Igor Rozhkov$^{3}$,\\ 
\bf Tatiana Batura${}^{3,4}$, 
Vladimir Ivanov${}^{5}$, 
Elena Tutubalina${}^{1,6,7}$\\
    	${}^{1}$HSE University 
    	${}^{2}$Huawei Noah's Ark lab, \\
    	${}^{3}$Lomonosov Moscow State University,
    	${}^{4}$Novosibirsk State University\\
    	${}^{5}$Innopolis University,
    	${}^{6}$Kazan Federal University,
        ${}^{7}$Sber AI, Russia\\}
\vspace{1cm}

\date{}

\begin{document}
\maketitle
\begin{abstract}

The RuNNE Shared Task approaches the problem of nested named entity recognition. The annotation schema is designed in such a way, that an entity may partially overlap or even be nested into another entity. This way, the named entity ``The Yermolova Theatre'' of type \textsc{organization} houses another entity ``Yermolova'' of type \textsc{person}. We adopt the Russian NEREL dataset \cite{loukachevitch-etal-2021-nerel} for the RuNNE Shared Task. NEREL comprises news texts written in the Russian language and collected from the Wikinews portal. The annotation schema includes 29  entity types. The nestedness of named entities in NEREL reaches up to six levels.  The RuNNE Shared Task explores two setups. (i) In the general setup all entities occur more or less with the same frequency.  (ii) In the  few-shot setup the majority of entity types occur often in the training set. However, some of the entity types are have lower frequency, being thus challenging to recognize. In the test set the frequency of all entity types is even.  

This paper reports on the results of the RuNNE Shared Task. Overall the shared task has received 156 submissions from nine teams. Half of the submissions outperform a straightforward BERT-based baseline in both setups. This paper overviews the shared task setup and discusses the submitted systems, discovering meaning insights for the problem of nested NER. The links to the evaluation platform and the data from the shared task are available in our \href{https://github.com/dialogue-evaluation/RuNNE}{github repository}. 

  
  \textbf{DOI:} 
\end{abstract}

\selectlanguage{russian}
\begin{center}
  \russiantitle{Соревнование RuNNE-2022: \\извлечение вложенных именованных сущностей}

  \medskip
        \textbf{Артемова Е.~Л.}${}^{1,2}$,
    	\textbf{Змеев М.~В.}${}^{1}$,
        \textbf{Лукашевич Н.~В.}${}^{3}$,
    	\textbf{Рожков И.C.}${}^{3}$, \\
    	\textbf{Батура Т.~В.}${}^{3,4}$,
    	\textbf{Иванов В.~В.}${}^{5}$,
    	\textbf{Тутубалина Е.~В.}${}^{2,6,7}$ \\
    	\medskip
    	${}^{1}$Национальный исследовательский университет Высшая школа экономики  \\
    	${}^{2}$Huawei Noah's Ark lab,\\
    	${}^{3}$Московский Государственный Университет им. М.В. Ломоносова\\
    	${}^{4}$Новосибирский государственный университет\\
    	${}^{5}$Иннополис,
    	${}^{6}$Казанский федеральный yниверситет, 
    	${}^{7}$Sber AI, Россия\\  
  
  \medskip
\end{center}

\begin{abstract}
Извлечение именованных сущностей – одна из самых востребованных на практике задач извлечения информации – предполагает поиск в тексте упоминаний имен, организаций, топонимов и других сущностей. Соревнование RuNNE посвящено задаче извлечения вложенных именованных сущностей. Разметка данных допускает следующие случаи: внутри одной именованной сущности находится другая именованная сущность. Так, например в сущность класса Organization ``Московский драматический театр имени М. Н. Ермоловой'' вложена сущность типа Person – ``М. Н. Ермоловой''. Соревнование проводится на материале корпуса NEREL\cite{loukachevitch-etal-2021-nerel}, собранного из новостных текстов WikiNews на русском языке. В корпусе NEREL представлено 29 классов различных сущностей, а глубина вложенности сущностей достигает 6 уровней разметки. В рамках соревнования RuNNE мы предлагаем участникам рассмотреть few shot постановку задачи. Задача предполагает извлечение вложенных именованных сущностей, В обучающем множестве большая часть типов именованных сущностей встречается достаточно часто, а некоторое количество специально отобранных типов – встречается всего несколько раз. В тестовом множестве все типы сущностей представлены одинаково.
В данной статье мы описываем соренование RuNNE, подводим его итоги и проводим сравнение решений, полученных от участников.

  \textbf{Ключевые слова:} извлеченние вложенных именованных сущностей, соревнование 
\end{abstract}
\selectlanguage{british}


\section{Introduction}
\label{intro}
Named entity recognition is one of the most popular tasks in natural language processing. It involves labeling mentions of personal names, organizations, toponyms and other entities in the text. In early works, only  ``flat'' named entities, in which internal named entities within a longer entity are not allowed,  were annotated and extracted from texts. However, the assumption that all entities are flat oversimplifies the task and leads to annotation collision. For example, consider such entities as the ``Ministry of Education of the Russian Federation'' or the ``State Duma of the Federal Assembly of the Russian Federation''. In this case, there are two options of flat labeling:  a) splitting into minimum spans (``Ministry of Education'' and ``Russian Federation''), or b) extracting  maximal spans. The latter case would miss the mention of the \textit{Russian Federation} as a separate entity. Both labeling approaches lead to loosing meaningful information pieces, which, in turn, can be useful for further processing, e.g. relation extraction. 

In recent years, recognition of nested named entities has received special attention. In case of nested NEs a longer named entity can contain internal named entities. In this research direction, multiple new  datasets were published \cite{plank2020dan,ringland2019nne,ruokolainen2019finnish}. The best performing methods for nested NER utilize specialized neural network architectures \cite{strakova2019neural,yu2020named,jue2020pyramid}. A recent dataset for nested NER in Russian, NEREL, is annotated with 29 entity types  \cite{loukachevitch-etal-2021-nerel}. 
 
The RuNNE shared task aims to popularize the nested NER task. To this end, we setup two sub-tasks based on the NEREL dataset: (i) the general nested NER evaluation; entities of more or less the same frequency are used (ii) the few-shot nested NER evaluation, where low-frequency entities are used. This paper (i) introduces the NEREL corpus (Section~\ref{section:data}), (ii) reports the results of the RuNNE Shared Task (Section~\ref{section:teams}) and (iii) compares submitted systems (Section~\ref{section:results}).


\section{Related NER Datasets and Shared Tasks}

Several datasets for named entity recognition in the Russian language are available, e.g. the Gareev's dataset \cite{Gareev}, Persons-1000 and Collection3 \cite{mozharova2016two,vlasova2014}, FactRuEval~\cite{starostin2016factrueval}, the Russian subset of the BSNLP dataset~\cite{piskorski2019second}, and RURED \cite{rured}. In particular, FactRuEval and BSNLP datasets were used for corresponding shared tasks. FactRuEval published in 2016 has been the only  Russian dataset annotated with nested named entities until recently. For example, a person object in FactRuEval could include name, surname, patronymic, and nickname spans.
The dataset is rather small with approximately 250 news texts annotated with 11,700 spans and 7,700 objects. This limits its effectiveness in building large neural network models.
Additional  descriptions and statistics of Russian NER datasets can be found in 
~\cite{loukachevitch-etal-2021-nerel}.



In recent years, the NLP community has held multiple shared tasks on named entity recognition, tackling different aspects of the task:
\begin{itemize}
    \item \textbf{Multilinguality} The series of SlavNER Shared Tasks \cite{piskorski-etal-2021-slav} focuses on six Slavic languages; the  MultiCoNER Shared Task~\cite{multiconer-report} features 11 world languages. XTREME~\cite{hu2020xtreme} and XGLUE~\cite{liang-etal-2020-xglue} data suites are collections of tasks and corresponding datasets for evaluation of zero-shot transfer capabilities of large multilingual models from English to other languages. XGLUE adopts CoNLL-2002 NER~\cite{conll2002ner} and CoNLL-2003 NER~\cite{conll2003ner} datasets covering four languages (English, German, Spanish, and Dutch) and four entity types (Person, Location, Organization, and Miscellaneous). XTREME uses Wikiann dataset~\cite{pan-etal-2017-cross}, where Person, Location and Organization entities were automatically annotated in Wikipedia pages in 40 languages.
    \item \textbf{Complexity} The MultiCoNER Shared Task~\cite{multiconer-report} was devoted to extraction of semantically ambiguous and complex entities, such as movie and book titles in short and low-context settings.
    \item \textbf{Applied domains} The WNUT initiative aims at developing NLP tools for noisy user-generated text. To this end, they run multiple shared tasks on NER and relation extraction from social media \cite{nguyen-etal-2020-wnut,chen-etal-2020-test}. PharmaCoNER is aimed at clinical and medical NER\cite{gonzalez-agirre-etal-2019-pharmaconer}.
    \item \textbf{Languages, other than English} The Dialogue evaluation campaign has supported two shared tasks for NER in Russian. FactRuEval dataset comprised news texts, Wikipedia pages and posts on social media, annotated according to a two-level schema. 
    The RuReBUS Shared task~\cite{ivanin2020rurebus} approached joint NER and relation extraction in Russian business-related documents.
\end{itemize}


\section{NEREL Dataset} \label{section:data}

The NEREL collection is developed for studying three levels of information extraction methods including named entity recognition, relation extraction and entity linking \cite{loukachevitch-etal-2021-nerel}.
Currently, NEREL is the largest Russian dataset annotated with entities and relations compared to the existing Russian datasets. NEREL comprises 29 named entity types and 49 relation types. At the time of writing, the dataset contains 56K named entities and 39K relations annotated in 900+ person-oriented news articles. NEREL is annotated with relations at three levels: within nested named entities, within and across sentences. Entity linking annotations leverage nested named entities, and each nested named entity can be linked to a separate Wikidata entity. 

\begin{table}[!h]

\centering
\small
\begin{tabularx}{\textwidth}{ >{\RaggedRight}p{0.15\linewidth} >{\RaggedRight}X >{\RaggedRight}X  }

\toprule

Entity  &  Example & Annotation  \\
\midrule

\multicolumn{3}{c}{\textbf{General entities}} \\
\midrule

\multirow{3}{*} {\textsc{profession} } 

& \textit{governor of California}
& [governor of [California]\textsubscript{\textsc{state\_or\_province}}]\textsubscript{\textsc{profession}}\\

& \textit{head of Gazprom}& [head of [Gazprom]\textsubscript{\textsc{organization}}]\textsubscript{\textsc{profession}}\\

\midrule

\multirow{3}{*}{\textsc{organization} }

&  \textit{physics department of Lomonosov Moscow State University}
& 
[physics department of [[Lomonosov]\textsubscript{\textsc{person}} [Moscow]\textsubscript{\textsc{city}} State University]\textsubscript{\textsc{org}}]\textsubscript{\textsc{org}} \\

& \textit{Russian Government} 
& [[Russian]\textsubscript{\textsc{country}} government]\textsubscript{\textsc{org}}\\ 
\midrule

\multirow{3}{*}{\textsc{nationality} }
& \textit{citizen of Russia}
& [citizen of [Russia]\textsubscript{\textsc{country}}]\textsubscript{\textsc{nationality}}\\

&  \textit{Russians} 
& [Russians]\textsubscript{\textsc{nationality}} \\ 

& \textit{Russian writer}
&[Russian]\textsubscript{\textsc{nationality}} [writer]\textsubscript{\textsc{profession}}\\

\midrule

\multirow{3}{*}{\textsc{law} }
& \textit{Yarovaya law}
& [[Yarovaya]\textsubscript{\textsc{person}} law]\textsubscript{\textsc{law}}\\

&  \textit {article 84 of the Constitution of Kyrgyzstan}
& [article [84]\textsubscript{\textsc{ordinal}} of the [[Constitution]\textsubscript{\textsc{law}} \textsubscript{\textsc{country}}]\textsubscript{\textsc{law}}]\textsubscript{\textsc{law}}\\

\midrule

\multirow{4}{*}{\textsc{crime} }
&  \textit{complicity in murder of VGTRK journalists} & [[complicity in murder]\textsubscript{\textsc{crime}} of [[VGTRK]\textsubscript{\textsc{org}} [journalists]\textsubscript{\textsc{profession}}] \textsubscript{\textsc{crime}} \\

&  \textit{armed attack on passers-by}&[[armed attack]\textsubscript{\textsc{crime}} on passers-by]\textsubscript{\textsc{crime}}\\

&  \textit{attempt to oust Erdogan}&[attempt to oust [Erdogan]\textsubscript{\textsc{person}}]\textsubscript{\textsc{crime}}\\

\midrule

\multirow{3}{*}{\textsc{product}}& \textit{Boeing-737 MAX}&[[Boeing]\textsubscript{\textsc{org}}-[737]\textsubscript{\textsc{number}} MAX]\textsubscript {\textsc{product}}\\
&  \textit{Apple Watch}&[Apple]\textsubscript{\textsc{org}} Watch]]\textsubscript{\textsc{product}}\\

\midrule
\multirow{5}{*}{\textsc{award}}& \textit{Merit for the Fatherland Order}&[Merit for the Fatherland Order]\textsubscript{\textsc{award}} \\

&   \textit{gold  of the Olympic Games}&[gold  of the [Olympic Games]\textsubscript{\textsc{event}}]\textsubscript{\textsc{award}}\\

&  \textit{champion of the Olympic Games}&[champion  of the [Olympic Games]\textsubscript{\textsc{event}}]\textsubscript{\textsc{award}}\\

&  \textit{Miss Russia-2017}&[Miss [Russia]\textsubscript{\textsc{country}}-[2017]\textsubscript{\textsc{date}}]\textsubscript{\textsc{award}}\\
& \textit{gold medal}&[gold medal]\textsubscript{\textsc{award}} \\
  
\midrule 

\multirow{4}{*}{\textsc{event}}& \textit{40th Moscow International Film Festival} & [[40th]\textsubscript{\textsc{ordinal}} [[Moscow]\textsubscript{\textsc{city}} International Film Festival]\textsubscript{\textsc{event}}]\textsubscript{\textsc{event}}\\

& \textit{UEFA champions league}
&[UEFA]\textsubscript{\textsc{org}} champions league]\textsubscript{\textsc{event}} \\
 
& \textit{Sochi-2014}
&[[Sochi]\textsubscript{\textsc{city}}-[2014]\textsubscript{\textsc{date}}]\textsubscript{\textsc{event}} \\

\midrule
\multicolumn{3}{c}{\textbf{Few-shot entities}} \\
\midrule

\multirow{2}{*}{\textsc{penalty}} & \textit{\$129 million fine}&[[\$129 million]\textsubscript{\textsc{money}} [fine]\textsubscript{\textsc{penalty}}]\textsubscript{\textsc{penalty}}\\

& \textit{imprisonment for 3 years}&[[imprisonment]\textsubscript{\textsc{penalty}} for [3 years]\textsubscript{\textsc{date}}]\textsubscript{\textsc{penalty}}\\

\midrule

\multirow{2}{*}{\textsc{disease}} & died from Covid & [[died]\textsubscript{\textsc{event}}  from [Covid]\textsubscript{\textsc{disease}}]\textsubscript{\textsc{event}} \\

& thyroid cancer & [thyroid [cancer]\textsubscript{\textsc{disease}}]\textsubscript{\textsc{disease}} \\

\midrule

\multirow{2}{*}{\textsc{work\_of\_art}} & the host of TV show ``Vzglyad'' & [the host of TV show [``Vzglyad'']\textsubscript{\textsc{work\_of\_art}}]\textsubscript{\textsc{profession}} \\

\bottomrule
\end{tabularx} 

\caption{Examples of annotating nested named entities of different types}
\label{tab:NEexamples}   
\end{table}

The NEREL corpus consists mainly of Russian Wikinews articles having size of 1--5 Kb, as such medium-sized texts are more convenient for annotation.  The BRAT tool~\cite{stenetorp2012brat} was used  for annotation. Three levels of annotation~-- named entities, relations, Wikidata links, -- were performed as independent subsequent passes. NEREL dataset is publicly available. It is subdivided into train, dev, and test parts, which were used in the RuNNE evaluation. The training part was specially reduced for few-shot NER evaluation.
Evaluation 
was carried out on the NEREL's dev and test sets. Table \ref{tab:NEexamples} contains  examples of nested named entities in the NEREL dataset.  Entity type statistics of the dataset used for the RuNNE competition are shown in Table \ref{tab:table_stats}. During the evaluation phase the NEREL github repository was temporarily closed.

\begin{table}[!h]

\centering
    \begin{tabular}{l r r r |l r r r}
\toprule
    Entity  & train & dev & test & Entity  & train & dev & test \\
\midrule

\textsc{profession} & 4593 & 860 & 854 & \textsc{ideology} & 300 & 36 & 43 \\
\textsc{person} & 4518 & 947 & 961 & \textsc{location} & 272 & 64 & 62 \\
\textsc{organization} & 4059 & 616 & 675 & \textsc{product} & 238 & 30 & 53 \\
\textsc{event} & 2879 & 707 & 690 & \textsc{crime} & 181 & 64 & 35 \\
\textsc{country} & 2521 & 355 & 456 & \textsc{money} & 171 & 29 & 43 \\
\textsc{date} & 2276 & 527 & 523 & \textsc{time} & 154 & 29 & 47 \\
\textsc{city} & 1102 & 208 & 239 & \textsc{district} & 98 & 18 & 25 \\
\textsc{number} & 1026 & 186 & 230 & \textsc{religion} & 94 & 9 & 24\\
\textsc{ordinal} & 565 & 102 & 107 & \textsc{percent} & 82 & 9 & 7 \\
\textsc{age} & 554 & 137 & 138 & \textsc{language} & 43 & 7 & 8 \\
\textsc{nationality} & 394 & 59 & 66 & \textsc{\textbf{disease}} & \textbf{32} & 117 & 57 \\
\textsc{law} & 392 & 77 & 62 & \textsc{\textbf{penalty}} & \textbf{32} & 57 & 18 \\
\textsc{facility} & 371 & 84 & 63 & \textsc{\textbf{work\_of\_art}} & \textbf{30} & 104 & 93 \\
\textsc{state\_or\_province} & 343 & 99 & 112 & \textsc{family} & 17 & 7 & 14 \\
\textsc{award} & 328 & 43 & 121 & \textbf{Total} & \textbf{27665} & \textbf{5587} & \textbf{5826} \\ 
\bottomrule
    \end{tabular}
    \caption{NEREL dataset statistics for the RuNNE competition.}
    \label{tab:table_stats}
\end{table}

\section{RuNNE Task}
The RuNNE competition offered two tasks: general nested named entity recognition and few-shot setting.  In the training set, most  named entity types occur quite often, and a certain number of specially selected types occur only a few times. In the test set, all entity types are represented equally.

As a quality metric in the RuNNE competition, macro averaging of the F1-measure is used in two versions:  by classes of known entities (general formulation of the problem of extracting nested named entities) and by classes of new named entities (few-shot formulation). Three entity types annotated in the NEREL dataset were selected for the few-shot setting: \textsc{disease, work\_of\_art, penalty}.

As a baseline we employed a RuBERT model \cite{kuratov2019adaptation} with a fully connected output layer, predicting each of the classes in the data, from where only the flat most internal entities were taken. For named entity encoding the IOBES (Inside-Out-Begin-End-Single)  scheme was used.

\section{Evaluation} 
\begin{table}[!ht]
\centering
\begin{tabular}{ll|l|c|c|l}
\toprule
\textbf{User} & \textbf{Team }  & \textbf{\# of runs} & \textbf{F1$_{full\_set}$} & \textbf{F1$_{few-shot}$} & System Summary\\\midrule

\textit{Baseline} & & - &0.674& 0.447& RuBERT\\
\midrule
\multicolumn{6}{c}{\textit{Participating Teams}}\\
\midrule
ksmith & Pullenti &44&\textbf{0.811}& \textbf{0.710} & Rule-based\\
abrosimov\_kirill &Saldon&20& \underline{0.741} & \underline{0.644} &Sodner model, labelling \\
fulstock & MSU-RCC &7&\underline{0.749}& \underline{0.604}&MRC model\\
svetlan & &23&0.607&\underline{0.572} &n/a\\
LIORI &  &6&0.653 & 0.433&n/a\\
botbot & & 8&0.460& 0.414 &n/a\\
bond005 & SibNN &20&\underline{0.743}& 0.404 &Siemese network, Viterbi alg.\\
Stud2022 & & 24&0.477& 0.395 &n/a\\
mojesty & &2 & 0.619& 0.172 &n/a\\
\bottomrule
\end{tabular}

\caption{Macro-averaged F-scores on the official test sets. The best results are in bold. Results above the baseline are underlined.}
\label{table:test_res}
\end{table}

\subsection{Participants' submissions} \label{section:teams}

We have received 156 submissions from nine teams. Table~\ref{table:test_res} presents with the final scores of the submitted systems. Four out of nine systems outperformed the baseline. Below, we give an overview of these approaches.


Team \textbf{Ksmith (Pullenti)} achieved the best results by using a rule-based approach. Customized rules were written for every entity type \cite{kozerenko2018semantic}. This team made the highest number of entries to the leaderboard.

Team \textbf{Fulstock (MSU RCC)} applied the Machine reading comprehension model (MRC) \cite{li2020unified}. The MRC model treats NER as a question-answering task. Entity types are translated into Russian. Next, their definitions are gathered from dictionaries in Russian and  used as questions. The MRC model comprises three binary classifiers over the output of the last hidden layer from RuBERT. The first classifier determines the starting position of a named entity. The second classifier decides about the end position of a named entity (perhaps another) of the same class. The third classifiers classifies, weather the start-end pairs are a single entity.  These classifiers are trained for each of the entity type. 

Team \textbf{Abrosimov\_Kirll  (Saldon)} used the span-based Sodner model \cite{li2021span} that can recognize both overlapped and discontinuous entities jointly. The model is based on the graph convolutional network architecture. It includes two steps.  First, entity fragments are recognized by traversing over all possible text spans. Second, relation classification is implemented to judge to detect overlapping or succession. The team used a pre-trained RuBERT model as contextualized encoder and the Natasha syntax parser.\footnote{\url{https://github.com/natasha/slovnet\#syntax}} Additionally, selected sentences for few-shot entity types were manually annotated.  

Team \textbf{Bond005 (SibNN)} fine-tuned a pre-trained BERT-based transformer two times. First, the contextualized encoder is fine-tuned as a Siamese neural network using the supervised contrastive learning loss. This step brings together embeddings of named entities of the same class and moves apart named entities of different classes. The second step fine-tunes  the contextualized encoder for sequence labeling using a task-specific loss function. Finally, the team applied the Viterbi algorithm to smooth prediction probabilities. The team defined the transition probabilities manually. 

\subsection{Results} \label{section:results}

The results (Table \ref{table:test_res}) indicate that:
\begin{enumerate}
    \item the conventional rule-based approach is capable of outperforming recent neural models. However, the rule-based approach requires careful configuration, leading to more attempts to submit to the leaderboard; 
    \item the MRC approach benefits from  additional textual descriptions and outperforms other learnable approaches;
    \item the few-shot setup benefits from additional manual labelling and hand-crafted rules. 
\end{enumerate}

\subsection{Detailed comparison of systems}
\label{subsection:analysis}

\begin{table}[!h]
    \centering
    \small
    
    \begin{tabular}{lrrrrrrrrrrrr}
    \toprule
Entity Type & baseln. & pullenti & abros. & MSU.& svetlan & LIORI & botbot & SibNN & Stud2022 & mojesty\\ 
    \midrule

\textsc{organization} & 23 & 21 & 28 & 21 & 20 & 18 & 17 & 29 & 14 & 24\\ 
\textsc{country} & 8 & 36 & 19 & 12 & 27 & 21 & 15 & 12 & 19 & 34\\ 
\textsc{work\_of\_art} & 22 & 3 & 20 & 16 & 1 & 16 & 2 & 24 & 1 & 23\\ 
\textsc{state\_or\_prov.} & 11 & 6 & 21 & 9 & 5 & 18 & 4 & 18 & 4 & 29\\ 
\textsc{event} & 10 & 9 & 17 & 8 & 20 & 7 & 5 & 13 & 15 & 8\\ 
\textsc{location} & 9 & 8 & 12 & 8 & 6 & 18 & 8 & 5 & 6 & 10\\ 
\textsc{facility} & 6 & 9 & 11 & 11 & 5 & 7 & 2 & 10 & 3 & 10\\ 
\textsc{nationality} & 8 & 2 & 6 & 7 & 8 & 10 & 2 & 9 & 5 & 7\\ 
\textsc{product} & 9 & 1 & 5 & 6 & - & 5 & 1 & 17 & - & 7\\ 
\textsc{city} & 6 & 5 & 7 & 5 & 6 & 9 & 3 & 5 & 3 & 11\\ 
\textsc{person} & 12 & 2 & 11 & 4 & 5 & 7 & 1 & 6 & 2 & 8\\ 
\textsc{family} & 9 & - & 10 & 6 & 2 & 7 & 1 & 8 & 1 & 7\\ 
\textsc{district} & 7 & 2 & 4 & 3 & 1 & 10 & 1 & 6 & 1 & 5\\ 
\textsc{award} & 8 & 1 & 9 & 3 & 2 & 3 & 1 & 4 & 1 & 5\\ 
\textsc{disease} & 2 & 3 & 1 & 3 & 5 & 3 & 2 & 6 & 4 & 7\\ 
\textsc{profession} & 5 & 1 & 5 & 5 & 1 & 3 & - & 5 & 1 & 6\\ 
 \midrule
Total span matches 
& 4401 & 4732 & 4948 & 4853 & 3511 & 4283 & 2363 & 4618 & 2361 & 4595 \\

\bottomrule
\end{tabular}

\caption{Number of test instances with incorrectly recognized entity type. Top-16 entities cover approximately 85\% of all mismatches. The last row shows the number of correctly recognized spans for each system. Symbol `-' means that all system's predictions were correct. All calculations were performed for the \textit{full\_set} test set.}
\label{table:entity_mismatch}
\end{table}

\begin{figure}
    \centering
    \includegraphics[width=0.8\textwidth]{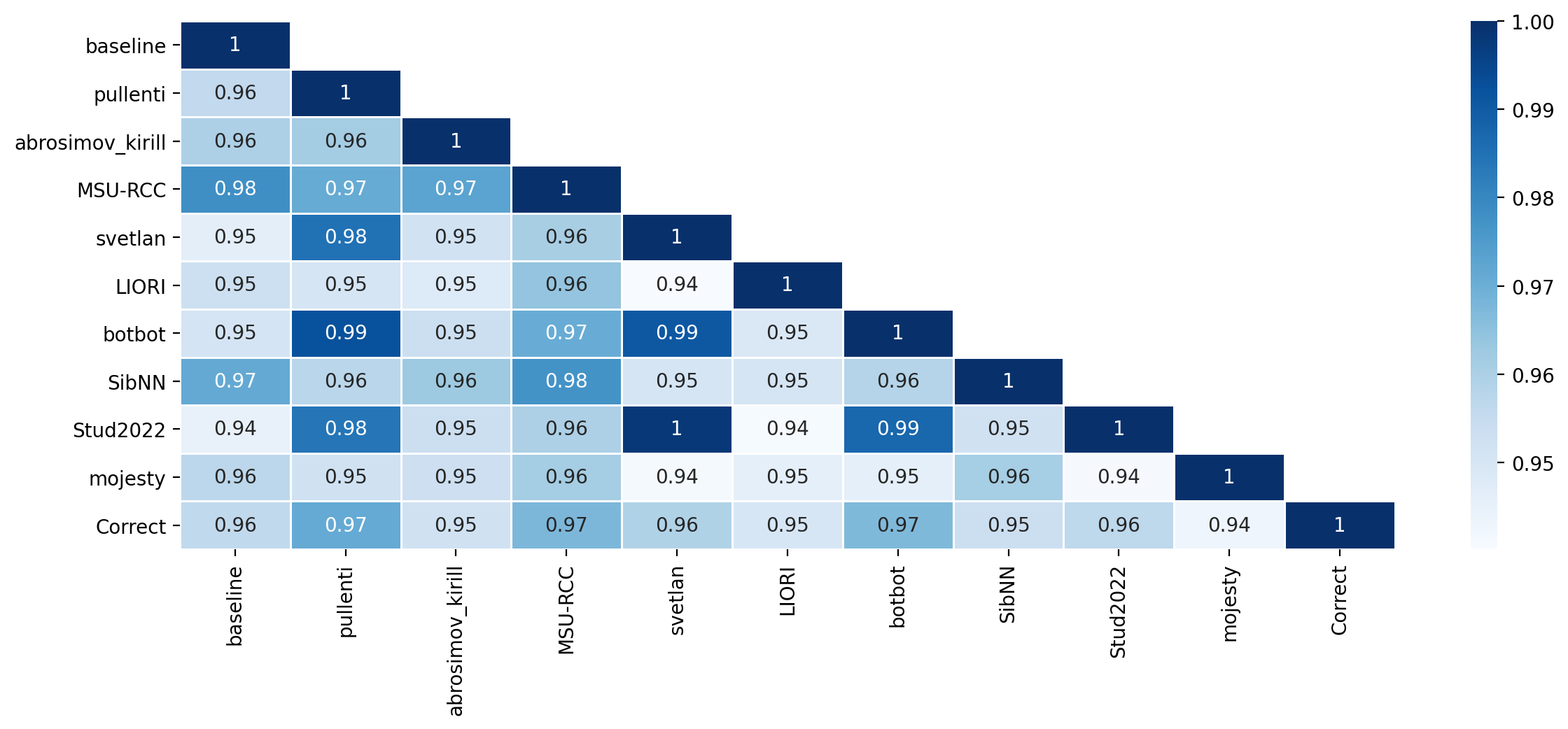}
    \caption{Agreement (Cohen's $\kappa$) among participating systems in annotating entity types. The ``Correct'' row corresponds to the test dataset annotations.}
    \label{fig:heatmap_triangle_aggr}
\end{figure}

We have compared solutions of participating systems in two aspects. First, we compare errors in entity type annotations given a span of a named entity was correctly detected. Second, we calculate pairwise agreements of systems' answers, again only for the annotations that have exact match of spans. Both evaluations are done on the test set. In Table \ref{table:entity_mismatch} we provide top entities for which systems give wrong entity type prediction. We found that around a half of errors come from six entity types: \textsc{organization, country, work\_of\_art, state\_or\_province, event}, and \textsc{location}. The most common mismatches are the following (correct entity is on the left of the `$\rightarrow$' sign): 
\begin{itemize}
\item \textsc{organization $\rightarrow$ facility / product / city}, 
\item \textsc{country $\rightarrow$ nationality / organization}, 
\item \textsc{work\_of\_art $\rightarrow$ organization / law / event}, 
\item \textsc{state\_or\_province $\rightarrow$ city / country / location}, 
\item \textsc{event $\rightarrow$ organization / crime / disease}, 
\item \textsc{location $\rightarrow$ state\_or\_province / city/ country}.
\end{itemize}

To assess pairwise agreements between systems, we build lists of spans that have exactly matching boundaries, but may have different type annotations. Then we calculate Cohen's $\kappa$ for each pair of systems (Fig. \ref{fig:heatmap_triangle_aggr}). Depending on a pair of systems, the number of matching spans varies between 1,294 and 5,826. Due to this difference, values in the Figure \ref{fig:heatmap_triangle_aggr} slightly deviate from the ranking of the systems. Three teams, showing mediocre performance  (\textbf{svetlan}, \textbf{botbot} and \textbf{Stud2022}) have highest agreement with each other. Also these three teams have the highest agreement with the \textbf{pullenti} team. This indicates that these systems learn named entity patterns, which can be described by rules, too. Other systems may gain more generalization abilities and thus are less correlated with the rule-based system. Finally, the main source of errors for all approaches may be attributed to ambiguous or complex cases.

\section{Conclusion}
In this paper we described the RuNNE Shared Task, aimed at extracting nested named entities from texts in Russian. We used NEREL \cite{loukachevitch-etal-2021-nerel} as a source dataset.  RuNNE comprises two sub-tasks: (i) the general nested NER evaluation and  (ii) the few-shot nested NER evaluation. Comparing submissions of the RuNNE participants, we found that the classic rule-based approach is capable to outperform recent neural models. However, it requires a special framework for writing rules  and iterative improvement of rules, leading to more attempts in leaderboard submits. In both subtasks, the best model among machine learning approaches applied without additional manual data labeling was  Machine Reading Comprehension model \cite{li2020unified}. 

\section*{Acknowledgments}
The project is supported by the Russian Science Foundation, grant \# 20-11-20166. The experiments were partially carried out on computational resources of HPC facilities at HSE University \cite{kostenetskiy2021hpc} and the shared research facilities of HPC computing resources at Lomonosov Moscow State University. Ekaterina Artemova was supported by the framework of the HSE University Basic Research Program.

\bibliography{anthology,dialogue}
\bibliographystyle{dialogue}



\end{document}